\documentclass[sigconf]{acmart}

\renewcommand\footnotetextcopyrightpermission[1]{} 
\usepackage{booktabs}
\usepackage{amsmath}
\usepackage{graphicx}

\title{The Prediction-Measurement Gap: Toward Meaning Representations as Scientific Instruments}

\author{Hubert Plisiecki}
\email{hplisiecki@gmail.com}
\orcid{0000-0002-5273-1716}
\affiliation{%
  \institution{IDEAS Research Institute}
  \city{Warsaw}
  \country{Poland}
}

\acmConference[Preprint]{}{}{}

\begin{document}

\begin{abstract}
Text embeddings have become central to computational social science and psychology, enabling scalable measurement of meaning and mixed-method inference. Yet most representation learning is optimized and evaluated for prediction and retrieval, yielding a prediction-measurement gap: representations that perform well as features may be poorly suited as scientific instruments. The paper argues that scientific meaning analysis motivates a distinct family of objectives - scientific usability - emphasizing geometric legibility, interpretability and traceability to linguistic evidence, robustness to non-semantic confounds, and compatibility with regression-style inference over semantic directions. Grounded in cognitive and neuro-psychological views of meaning, the paper assesses static word embeddings and contextual transformer representations against these requirements: static spaces remain attractive for transparent measurement, whereas contextual spaces offer richer semantics but entangle meaning with other signals and exhibit geometric and interpretability issues that complicate inference. The paper then outlines a course-setting agenda around (i) geometry-first design for gradients and abstraction, including hierarchy-aware spaces constrained by psychologically privileged levels; (ii) invertible post-hoc transformations that recondition embedding geometry and reduce nuisance influence; and (iii) meaning atlases and measurement-oriented evaluation protocols for reliable and traceable semantic inference. As the field debates the limits of scale-first progress, measurement-ready representations offer a principled new frontier.
\end{abstract}



\maketitle


\section{Introduction}
Text representations have been a focus of computational science for some time, allowing scientists to extract quantitative, or mixed-methods conclusions from data that would otherwise be analyzed only on the qualitative level. Research has shown that they can be used to recover the relational structure of culture (e.g. social constructs of class; \cite{kozlowski_geometry_2019}), uncover society-level stereotypes related to gender and ethnicity \cite{garg_word_2018}, as well as tackle the questions of whether psychometric questionnaires measure the same concepts \cite{wulff_semantic_2025}, and understand individual differences in what text people produce \cite{plisiecki_measuring_2025}. This line of research treats text embeddings as a way to quantify the structure of meaning understood as patterned relations among concepts that can be either studied directly, or applied to study specific linguistic artifacts like psychometric questionnaires \cite{arseniev-koehler_theoretical_2024,kozlowski_geometry_2019,wulff_semantic_2025}. 

It is therefore worth asking whether NLP optimization criteria align with the needs of social science. This mismatch is referred to here as the prediction-measurement gap. As the recent progress in language modeling has been mostly focused on scaling transformer architectures and optimizing training for efficient predictive performance \cite{hoffmann_training_2022,kaplan_scaling_2020}, alongside the research on adapting foundation models for specific tasks \cite{bommasani_opportunities_2021}, an injection of an external social science perspective might be useful in setting new research pathways. While the current NLP paradigm has been remarkably successful, it has also led to the status quo in which representation quality is mostly assessed via predictive or retrieval benchmarks, rather than through criteria that matter when representations are used as scientific instruments. Large embedding benchmarks already highlight that “good” representations are task-dependent and that no single model dominates across use cases \cite{muennighoff_mteb_2022}, suggesting that additional objective families may be warranted.

The paper argues that scientific meaning analysis introduces such an objective family. When embeddings are used for measurement, explanation, and theory building, researchers need representations that support interpretability and traceability to linguistic evidence, compatibility with regression-style inference over semantic directions, and increased sensitivity to semantic information, rather than syntax and punctuation. These properties sit on the far side of the prediction–measurement gap. The contribution of this paper is therefore to operationalize the criteria for constructing language modeling approaches that are more useful to social science, as well as outline a research agenda for developing meaning representations and evaluation protocols that complement the prevailing NLP paradigm, thereby benefiting computational social science while also brodening the representation learning frontier. he paper proceeds by reviewing current uses of embeddings in social science (Section 2), formalizing the success criteria (Section 3), grounding them in cognitive and neuropsychological research (Section 4), assessing existing paradigms against them (Section 5), and sketching initial directions for a research agenda (Section 6).

\section{Text Representations in Social Science and Psychology}

Text representations are currently used across social sciences and psychology as both analytic instruments and measurement tools, making it possible to quantify meaning instead of relying on more subjective qualitative frameworks, as well as process amounts of data otherwise intractable through manual coding.

\subsection{Semantic Scaling and Society-Level Constructs}

One line of work focuses on society-level meaning structures, which can be extracted from large text corpora with the use of word embeddings. For example, embeddings have been used to map the meanings of class and their organization in semantic space and relate it to sociological theories \cite{kozlowski_geometry_2019}, as well as to formalize the idea that they can serve as a computational model of cultural learning by analyzing how they encode the meaning of being overweight across time \cite{arseniev-koehler_machine_2022}. A related strain of research used the geometry of word embedding spaces to quantify how the meaning of gender and ethnicity changed across time pinpointing similarities with theories of how these stereotypes evolved culturally \cite{garg_word_2018}. In these applications researchers used word embedding models to create interpretable linear lexical dimensions from predefined concept dictionaries - a technique formalized further by Watanabe, and named Latent Semantic Scaling \cite{watanabe_latent_2021}. These approaches have been proven to recover word embedding based scores for features such as age, danger, size, weight, wetness and many others \cite{grand_semantic_2022}.

\subsection{Embeddings for Psychometric Measurement}

A second line of work contains research that uses text representations to enhance psychometric measurement. Notably, recent work by Wulff and Mata \cite{wulff_semantic_2025} showed that transformer based text representations can be used to tackle the problem of the jingle-jangle fallacies which are a prevalent issue in psychology - namely that some separate psychometric scales measure the same underlying latent construct, while others despite being labeled similar capture unique phenomena. Based on a dataset including 459 scales they showed that, given a specifically finetuned encoder, through a cosine similarity analysis of questionnaire items vs. construct labels they were able to produce some face-valid results, but the cut-off threshold for questionnaires belonging to either of the problematic groups remained largely arbitrary. Creating proper numerical representations of questionnaires can also be useful in generative psychometrics where they are used to analyze the latent structure of items generated with the use of Large Language Models with some recent work extending the methods of extracting them from transformer models using exploratory graph analysis \cite{garrido_estimating_2025}.

\subsection{Interpretable Semantic Dimensions and Mixed-Method Inference}

A third line uses text representation geometry to extract interpretable semantic dimensions that enable mixed-methods interpretation. Most of the work in this category is related to mapping semantic change, which relates to how words and concepts change their meaning in time. These approaches are widely dominated by static embedding approaches that allow post-hoc qualitative interpretability via nearest neighbors \cite{hamilton_diachronic_2016}, however some more recent works extended them to contextual embeddings by comparing the probability distributions of substitutes of masked terms \cite{card_substitution-based_2023}. While these methods usually require sizable datasets for proper inference, a recent work extended the same interpretability logic to the measurement of individual differences by proposing the Supervised Semantic Differential (SSD) framework that works well in low-data environments such as psychological studies \cite{plisiecki_measuring_2025}. SSD aggregates pretrained embeddings from local contexts of a target concept in participant writings into person-level vectors and estimates a semantic gradient via regression on an external variable (e.g., a trait score). The gradient can then be interpreted using clusters of nearest neighbours, rearing both social science friendly statistical results from the regression as well as mixed-methods interpretability.

\subsection{Takeaways for Representation Design in Social Science}

Across these categories numerical text representations reveal themselves as extremely useful to social science, and psychological scientific inquiry. In particular, static word embeddings, despite no longer being the dominant representation paradigm, remain attractive because they support relatively transparent geometric operations (e.g., projections and linear semantic directions) and because their spaces often exhibit a concept-level organization that aligns with mixed-method interpretation and statistical inference goals. Even in the second category, which currently relies heavily on contextual embeddings, progress would likely come from representations whose geometry more cleanly reflects the structure of the constructs being measured, as well as allows for deeper explainability thereby assessing construct validity. Notably, while a parallel line of computational social science uses text representations for predictive modeling (e.g., sentiment trends in publicly expressed emotions; \cite{wang_global_2022}), this paper focuses on the representation properties required when embeddings are used as measurement instruments relevant to the broader prediction–measurement gap. That said, predictive applications also stand to benefit from representations with better-conditioned semantic geometry, since they can improve robustness, comparability across settings, and interpretability of the model's behavior when needed. These takeaways motivate the formalization of the success criteria presented in the next section.

\section{Core Success Criteria for Meaning Representations}
To narrow the prediction-measurement gap the paper formalizes a set of success criteria for meaning representations to make them more aligned with the needs of social sciences.

\subsection{Geometric Legibility}
A usable meaning representation should have a geometry in which distances and directions behave in stable and interpretable ways as accessible through simple operations (similarity, offsets, projections) rather than encoded in anisotropic or nonlinear structures. A possible operationalization is stability of cosine similarity across corpus subsamples, with a small number of linear components capturing semantically interpretable variance in targeted semantic dimensions.

\subsection{Interpretability and Traceability to Linguistic Evidence}
Arbitrary latent vectors should be traceable to interpretable text outputs for instance, via nearest neighbors, thereby enabling qualitative inference beyond quantitative results. A candidate measure is whether blind annotation of a semantic direction's top-k neighbors or other linguistic artifacts yields adequate inter-rater agreement on the construct it represents.

\subsection{Semantic Sensitivity with Robustness to Non-Semantic Confounds}
Representations should encode rich semantic information while remaining robust to contaminating signals such as token frequency, punctuation, or tokenization artifacts. This can be assessed via a sensitivity contrast: the ratio of similarity across meaning-preserving surface variants to similarity across surface-similar but meaning-distinct strings.

\subsection{Cognitive Plausibility as a Design Constraint}
Compatibility with how constructs are organized in human cognition may improve both representation quality and adoption among psychologists, though this criterion should not override reliability requirements. A potential operationalization is whether similarity structure correlates more strongly with typicality ratings at the basic level of categorization than at superordinate or subordinate levels, as elaborated in Section 4.

\subsection{Summary and Scope}
These criteria formalize what would make text representations more useful as scientific instruments, grounded in the social science applications reviewed above. They are not exhaustive, but serve as a signpost for further research. Before examining to what extent existing text representation approaches satisfy them, the following section reviews cognitive and neuro-psychological evidence on how meaning is organized in the human mind, both to contextualize the first three criteria and to extend the fourth.

\section{Conceptual Grounding from Cognitive and Neuro-Psychological Research}

Within the psychological paradigm, meaning has classically been treated as a measurable property. A good example here is the work of Osgood and associates presented in their book "The Measurement of Meaning" \cite{osgood_measurement_1957}, where they used Likert-based categorical scales across many dichotomies (cold vs. warm, big vs. small) to measure meaning of multiple words and then extract the principal components of meaning (evaluation, potency and activity, - roughly corresponding to valence, arousal, and dominance). Their work inspired a large part of current psycholinguistic research, and is conceptually congruent with the modern distributional approaches to encoding meaning, such as word embeddings.

\subsection{Cognitive Research on Meaning}
Cognitive research further suggests that semantic structure is graded, prototype-centered, and often organized across levels of abstraction. Notably, Conceptual Spaces theory treats concepts as regions in an interpretable geometric space, with typical instances near region centers and similarity reflected by distance \cite{gardenfors_conceptual_2000,gardenfors_geometry_2014}. This is supported by the fact that simple projections in word embedding spaces can recover context-dependent human judgments of object features, functioning like “mental scales” (Grand et al., 2022). Categorization research adds an important constraint for any hierarchy-aware representation: taxonomies are not psychologically uniform. In Rosch’s basic-level account, categories reflect bundles of correlated attributes in the world, and the “basic level” (e.g., chair) is privileged because it best balances informativeness with cognitive economy \cite{rosch_basic_1976,mervis_categorization_1981}. Superordinate categories (e.g., furniture) have high within-category variance in feature bundles due to its members sharing few concrete attributes making them harder to summarize with a single prototype. Subordinate categories (e.g., kitchen chair) reduce variance further but offer diminishing returns in distinctiveness relative to added specificity. Basic-level categories sit at the sweet spot where within-category feature bundles are coherent (lower variance) while between-category bundles are maximally separable (higher contrast), making them easiest to name, recognize, and learn.

\subsection{Neuro-Psychological Research on Meaning}

Neuro-psychological evidence shows that semantic information is supported by specialized brain systems, lending credence to the task of separating it from other linguistic cues. The Controlled Semantic Cognition framework argues that semantic knowledge is represented as distributed feature information across modality-linked systems (“spokes”) and integrated by a trans-modal hub centered in the anterior temporal lobes, enabling generalizable concept representations that support inference beyond any single sensory or linguistic cue \cite{ralph_neural_2017}. In parallel, integrative neural modeling of language shows that large language models can align closely with neural and behavioral responses during comprehension, and that this alignment is correlated with next-token prediction performance \cite{schrimpf_neural_2021}. However, the alignment with brain activation does not equate pure semantic representation. Text comprehension recruits multiple interacting signals (syntax, form context), likely reflected in distinct neural patterns alongside semantic representations. Thus, better next-token prediction does not directly imply better semantic representations for scientific measurement.

\subsection{Summary of Neuro-Cognitive Evidence}

In summary, cognitive and neuro-psychological literature supports a view of meaning representations as geometrically legible and prototype-structured, organized across constrained levels of abstraction, and separable from other linguistic signals. It is important to mention that the literature presented here is by no means exhaustive, but selected with the goal of informing text representation research. The next section will assess to what extent current text representation methods fit the scope set out by both research needs and grounding in neuro-cognitive literature.

\section{Current Text Representation Methods}

To assess existing representation paradigms against our success criteria, we focus on what semantic structure each paradigm makes accessible for measurement and mixed-method inference to better understand where they sit in the prediction-measurement gap.

\subsection{Static Embeddings}
Static embeddings remain an unusually good “default substrate” for scientific meaning analysis because they offer a transparent compromise between expressive power and measurement usability. By representing meaning at the word type level, they make it easy to construct concept dictionaries, compute centroids, estimate linear semantic directions, and interpret results by directly inspecting nearest neighbors, which constitute operations that align naturally with the regression-and-projection logic behind mixed-method frameworks such as semantic scaling and SSD-like approaches \cite{garg_word_2018,grand_semantic_2022,kozlowski_geometry_2019}. Just as importantly, the word-as-unit design tends to suppress many nuisance signals that complicate contextual representations (e.g., tokenization artifacts, casing, punctuation, position-dependent syntax), making the resulting geometry closer to a “semantic measurement space” than to a general-purpose sequence model.

At the same time, static embeddings only approximate the structure of meaning that social scientists often want. Their core limitation is that a single vector conflates multiple senses, so “meaning” is partly an average over contexts rather than a context-sensitive state. That said, this limitation is not purely fatal: evidence suggests that polysemy is often linearly superposed inside static embeddings, implying that sense structure can sometimes be recovered with additional modeling rather than requiring an immediate jump to fully contextual representations \cite{arora_linear_2018}. A second limitation is geometric conditioning: even static spaces can develop dominant directions and concentration effects that distort similarity and direction-based inference, but these distortions are often addressable with simple post-processing such as removing top components or whitening, which improves the legibility of distances and projections without changing the underlying vocabulary-based interpretability \cite{mimno_strange_2017,mu_all-but--top_2017,su_whitening_2021}.

Finally, the stability of static embeddings matters, but mainly insofar as downstream scientific use relies on local neighborhoods or cross-corpus comparability. Stability can vary with algorithmic choices and language properties, including morphology, which can fragment evidence across many surface forms and reduce reliability of word neighborhoods \cite{burdick_analyzing_2020,wendlandt_factors_2018}. For measurement-centric workflows that aggregate many tokens and focus on estimated semantic directions (as in SSD), further interpreting them through clustering, these instabilities are often attenuated, but they still motivate evaluating representations by the reliability of extracted semantic measurements (e.g., resampling stability of semantic gradients) rather than treating “embedding stability” as an end in itself \cite{du_measuring_2023}. Taken together, static embeddings satisfy many of our criteria by default (especially geometric legibility and interpretability), while their shortcomings (polysemy, conditioning, and some stability issues) point directly to principled extensions (sense-aware structure, geometry regularization, and hierarchy-aware manifolds) rather than to abandoning measurement-ready spaces altogether.

\subsection{Contextual Embeddings}

On the other hand, contextual language models substantially expand what a representation can capture: instead of assigning one vector per word type, they produce context-dependent token representations that separate polysemous meanings and encode relational information. This makes them highly effective for predictive tasks and for settings where the meaning of an expression depends strongly on local context. However, when evaluated against measurement-centric success criteria, contextual representations pose a different problem than static embeddings: they do not merely add semantic richness, they also change what kind of object an embedding is—shifting from a relatively stable lexical coordinate system to a layered, entangled representation optimized for next-token prediction.

A central challenge is that contextual representations are not “semantic-only” spaces. Across layers, they mix surface cues, syntax, and longer-range semantic dependencies, and different depths specialize in different kinds of linguistic information \cite{jawahar_what_2019,peters_dissecting_2018}. For scientific meaning analysis this creates an immediate ambiguity: which layer is the appropriate substrate for measurement, and under what justification? Even if one selects a layer, the geometry of contextual spaces is often poorly conditioned for direct similarity- or direction-based inference: anisotropy and other global distortions can make cosine similarity less reflective of semantics, and contextualized instances of the same word form high-dimensional clouds rather than collapsing into a small set of discrete sense prototypes \cite{ethayarajh_how_2019,li_sentence_2020}. Recent analyses further suggest that poor “semantic isometry” is not reducible to anisotropy alone but is strongly influenced by non-semantic biases such as frequency, casing, punctuation, and subword artifacts, which can dominate the representation geometry unless explicitly controlled \cite{fuster_baggetto_is_2022}. In short, contextual models often encode rich semantic information, but they do not necessarily present it in a form that is disentangled from other information and legible for the kinds of linear inference and traceable interpretation used in mixed-method workflows. These issues are amplified in the low-data regimes common in psychology and much of CSS, where instability in extracted semantic measurements complicates reproducibility and the interpretation of results \cite{du_measuring_2023,mosbach_stability_2021,risch_bagging_2020}.

Finally, recent work reveals that contextual representations preserve far more information about the input than is often assumed, including signals unrelated to semantics, and may even be invertible given sufficient access assumptions \cite{morris_text_2023,nikolaou_language_2025}. However, such invertibility applies to internal states generated by actual prompts, and does not extend to derived representations produced by aggregation or projection (e.g., semantic directions), which need not correspond to any valid point on the model’s representation manifold. Together, these considerations suggest that contextual embeddings should not be treated as a drop-in replacement for static semantic spaces in measurement pipelines; rather, they highlight the need for representation learning approaches explicitly designed for scientific usability, motivating the agenda set forward within the next section.

\section{Research Agenda and Discussion}
The analysis above suggests that the core bottleneck is not whether modern models contain semantic information, but whether representation learning objectives make that information usable for scientific inference. A pragmatic agenda, therefore, is to treat “meaning representations for science” as an explicit design target rather than an emergent byproduct of next-token prediction \cite{schrimpf_neural_2021}. Several research directions can be drafted to that end, framed as hypotheses about where progress may come from rather than as definitive solutions.

\subsection{Geometry-First Design for Semantic Usability}
A first direction concerns the diversification of geometric and statistical objectives: social-scientific analyses routinely require both continuous semantic gradients (suited to Euclidean-style linear operations) and structured abstraction that simplify interpretation (often better captured by hierarchy-aware geometries). Work on hyperbolic representations and emerging “hyperbolic LLM” variants already indicates that manifold choice can encode inductive biases about hierarchy efficiently \cite{nickel_poincare_2017,patil_hyperbolic_2025}, while cognitive literature may inform their further development: human taxonomies are not psychologically uniform, with a privileged basic level of abstraction where categories are maximally distinctive and nameable \cite{rosch_basic_1976}. Translating this into representation design suggests a potentially useful refinement of hierarchy-aware spaces: rather than mapping abstraction monotonically to radius, one could encourage a “basic-level band” (e.g., via a radius prior or warping/mixture-of-curvature design) where representational capacity and separability are highest, reserving the center for coarse superordinates and the outer region for fine-grained subtypes or senses — a design directly motivated by the cognitive plausibility criterion (Section 3.4), and testable via its operationalization: whether similarity structure aligns more strongly with typicality ratings at the basic level than at adjacent levels.

\subsection{Invertible Post-Hoc Transformations for Embedding Spaces}
A second direction concerns the reconditioning of existing embedding spaces: post-hoc, invertible transformations are likely to remain important even when new representation geometries are proposed, because much of today's infrastructure will continue to be transformer-based (at least for the foreseeable future). Work on geometric conditioning already demonstrates that invertible mappings can reorganize embedding geometry to make similarity more semantically aligned without retraining the base model \cite{li_sentence_2020}, while other analyses emphasize that anisotropy alone is an incomplete diagnosis and that non-semantic biases (frequency, casing, punctuation, subword artifacts) can dominate geometry unless actively controlled \cite{fuster_baggetto_is_2022} — precisely the failure mode captured by the sensitivity contrast in Section 3.3. This points to an agenda that combines conditioning with explicit robustness to non-semantic information by treating interpretability as an optimization and evaluation target. Recent advances in simulation-free training of continuous normalizing flows via flow matching and optimal-transport variants provide a scalable toolbox for learning such invertible transformations \cite{lipman_flow_2022,tong_improving_2023}, which is attractive precisely because it preserves information while reshaping geometry, making it useful when the goal is to make semantic structure accessible without assuming representations are already “semantic summaries”. Whether such flow-based methods will be readily adoptable in social-scientific workflows remains an open practical question, given typical constraints favoring simple and transparent statistical inference methods.

\subsection{Meaning Atlases and Measurement-Oriented Evaluation}
A third direction suggests that closing the prediction-measurement gap in the case of contextual models may require moving beyond "one vector per word", as the relevant semantic content is distributed across contexts and mixed with non-semantic signals, making single vectors poor semantic units for measurement. Interpretability in such settings might therefore require constructing explicit, human-readable anchors—e.g., phrases, definitions, prototypes, and (where helpful) sense-like units—that act as stable reference points for nearest-neighbor explanation and directional interpretation \cite{amrami_towards_2019,scarlini_more_2020}. A longer-term goal could be to build meaning atlases for transformer-based models: structured dictionaries of interpretable anchors — potentially including sense-extracted units derived from contextual substitutes — designed specifically to support measurement workflows. Atlas construction is, however, a hard coverage problem and may only be tractable in domain- or construct-specific "atlas fragments" rather than as a universal map. More generally, any agenda of this kind requires evaluation that matches the intended epistemic use. Existing embedding benchmarks assess representation quality via predictive or retrieval performance, but do not evaluate what matters when embeddings are used for inference and theory building \cite{muennighoff_mteb_2022}. A measurement-oriented evaluation suite would instead operationalize the criteria of Section 3 directly: geometric legibility via cosine similarity stability and linear component structure, interpretability via inter-rater agreement on nearest-neighbor annotations of semantic directions, and semantic sensitivity via the surface-variant contrast ratio. Such a suite would complement existing benchmarks with criteria suited to using text representations as scientific instruments.

\section{Summary and Conclusions}
This paper frames scientific meaning analysis as a distinct objective family and uses it to articulate the difference between representations optimized for predictive utility and those usable as scientific instruments for measurement, explanation, and inference. The agenda above is best read as course-setting: it invites the NLP community to evaluate meaning representations not only by downstream performance, but also by the semantic structure they make accessible and interpretable. A realistic near-term outcome is not a universal semantic space, but a toolbox: geometry conditioning methods, interpretable anchor constructions, domain-appropriate meaning atlases, and measurement-oriented benchmark protocols. Longer-term work may explore manifold design and hybrid geometries that better match the structure of semantic phenomena (hierarchies, prototypes, continuous dimensions), informed by cognitive constraints. This agenda is particularly relevant as the field debates the limits of scale-first progress \cite{chen_revisiting_2025}: measurement-ready meaning spaces offer a principled new frontier that neither competes with nor merely serves predictive NLP, but extends it toward semantics that is legible for social science.

\bibliographystyle{ACM-Reference-Format}
\bibliography{references}

\end{document}